\documentclass[final]{cvpr}

\usepackage{times}
\usepackage{epsfig}
\usepackage{graphicx}
\usepackage{amsmath}
\usepackage{amssymb}

\usepackage{tabularx}
\usepackage{booktabs}
\usepackage{graphicx}
\usepackage{animate}
\usepackage{multirow}
\usepackage{subcaption}
\usepackage{diagbox}
\usepackage[table,xcdraw,dvipsnames]{xcolor}
\usepackage{cite}

\usepackage[pagebackref=false,breaklinks=true,colorlinks,bookmarks=false]{hyperref}

\def\cvprPaperID{} 
\def\confYear{CVPR 2021}

\begin{document}

\title{\textit{Win-Fail} Action Recognition}

\author{Paritosh Parmar \hspace{1cm} Brendan Morris\\
University of Nevada, Las Vegas\\
{\tt\small parmap1@unlv.nevada.edu; brendan.morris@unlv.edu}
}

\maketitle

\begin{abstract}
Current video/action understanding systems have demonstrated impressive performance on large recognition tasks.  However, they might be limiting themselves to learning to recognize spatiotemporal patterns, rather than attempting to thoroughly understand the actions. To spur progress in the direction of a truer, deeper understanding of videos, we introduce the task of win-fail action recognition –- differentiating between successful and failed attempts at various activities.  We introduce a first of its kind paired win-fail action understanding dataset with samples from the following domains: ``General Stunts,'' ``Internet Wins-Fails,'' ``Trick Shots,'' \& ``Party Games.'' Unlike existing action recognition datasets, intra-class variation is high making the task challenging, yet feasible.  We systematically analyze the characteristics of the win-fail task/dataset with prototypical action recognition networks and a novel video retrieval task.  While current action recognition methods work well on our  task/dataset, they still leave a large gap to achieve high performance. We hope to motivate more work towards the true understanding of actions/videos. Dataset will be available from \url{https://github.com/ParitoshParmar/Win-Fail-Action-Recognition}.
\end{abstract}
\section{Introduction}
\begin{figure}
\centering
\footnotesize
\begin{tabular}{@{}cc|cc@{}}
\multicolumn{2}{c}{\begin{tabular}[c]{@{}c@{}}\textbf{Typical Action}\\ \textbf{Recognition Dataset}\end{tabular}} & \multicolumn{2}{|c}{\begin{tabular}[c]{@{}c@{}}\textbf{Ours Action} \\ \textbf{Understanding Dataset}\end{tabular}} \\ \midrule

\textbf{BB}                                     & \textbf{TS}                                     
& \textbf{Wins}                                           & \textbf{Fails}                                           \\

\animategraphics[loop,autoplay,poster=1,height=0.1\textwidth]{3}{Figs/Videos/Extracted_Samples_UCF101/Basketball/1/img_}{1}{16}   & \animategraphics[loop,autoplay,poster=1,height=0.1\textwidth]{3}{Figs/Videos/Extracted_Samples_UCF101/TennisSwing/1/img_}{1}{16} 
& \animategraphics[loop,autoplay,poster=1,height=0.1\textwidth]{3}{Figs/Videos/Extracted_Samples_sq/awefail/1/g/img_}{1}{16}   & \animategraphics[loop,autoplay,poster=1,height=0.1\textwidth]{3}{Figs/Videos/Extracted_Samples_sq/awefail/1/b/img_}{1}{16}    \\ 

\animategraphics[loop,autoplay,poster=1,height=0.1\textwidth]{3}{Figs/Videos/Extracted_Samples_UCF101/Basketball/3/img_}{1}{16}   & \animategraphics[loop,autoplay,poster=1,height=0.1\textwidth]{3}{Figs/Videos/Extracted_Samples_UCF101/TennisSwing/3/img_}{1}{16} 
& \animategraphics[loop,autoplay,poster=1,height=0.1\textwidth]{3}{Figs/Videos/Extracted_Samples_sq/trickshot/1/g/img_}{1}{16}   & \animategraphics[loop,autoplay,poster=1,height=0.1\textwidth]{3}{Figs/Videos/Extracted_Samples_sq/trickshot/1/b/img_}{1}{16}    \\ 

\animategraphics[loop,autoplay,poster=1,height=0.1\textwidth]{3}{Figs/Videos/Extracted_Samples_UCF101/Basketball/4/img_}{1}{16}   & \animategraphics[loop,autoplay,poster=1,height=0.1\textwidth]{3}{Figs/Videos/Extracted_Samples_UCF101/TennisSwing/4/img_}{1}{16} 
& \animategraphics[loop,autoplay,poster=1,height=0.1\textwidth]{3}{Figs/Videos/Extracted_Samples_sq/partygames/1/g/img_}{1}{16}   & \animategraphics[loop,autoplay,poster=1,height=0.1\textwidth]{3}{Figs/Videos/Extracted_Samples_sq/partygames/1/b/img_}{1}{16}    \\ 

\animategraphics[loop,autoplay,poster=1,height=0.1\textwidth]{3}{Figs/Videos/Extracted_Samples_UCF101/Basketball/5/img_}{1}{16}   & \animategraphics[loop,autoplay,poster=1,height=0.1\textwidth]{3}{Figs/Videos/Extracted_Samples_UCF101/TennisSwing/5/img_}{1}{16} 
& \animategraphics[loop,autoplay,poster=1,height=0.1\textwidth]{3}{Figs/Videos/Extracted_Samples_sq/general/1/g/img_}{1}{16}   & \animategraphics[loop,autoplay,poster=1,height=0.1\textwidth]{3}{Figs/Videos/Extracted_Samples_sq/general/1/b/img_}{1}{16}\\ 
\end{tabular}
\caption{\textbf{Illustration of intra-class variance} (along columns, not rows) in a typical action recognition dataset vs. in our action understanding dataset. (Left) Samples from two randomly chosen classes, Basketball (BB) and TennisSwing (TS), from a typical action recognition dataset. (Right) Samples from ours newly compiled dataset, which has two action classes: \textit{wins} and \textit{fails}. As can be seen, the typical action recognition dataset does not have much intra-class variance, because of which action recognition is reduced to a pattern recognition problem. \textit{Please view in an Adobe Reader to play videos.}}
\label{fig:ar_vs_ar}
\end{figure}
Action recognition, which can be defined as the task of identifying various action classes in videos, has thus far been used as a representative task for video understanding. Video action recognition involves the processing of spatiotemporal data, and extracting low-dimensional spatiotemporal signatures from video volumes. Based on these signatures, probabilities of action classes are determined.

We make two observations regarding the task of action recognition. Firstly, while current action recognition datasets, like UCF101, HMDB51, Kinetics, Sports1M, \etc, have focused on increasing their dataset sizes and covering a larger number of classes, samples in those datasets exhibit low intra-action-class variance in their spatiotemporal signatures. For example, all of the samples from action class BasketBall contain identical spatiotemporal signatures, like people holding a basketball with their hands, and trying to throw it into the basket (refer to Fig. \ref{fig:ar_vs_ar}); or as another example, all the samples from action class TennisSwing contain people holding a tennis racquet in their hand, and moving their arm. As a result, action recognition has, so far, been limited to cases where spatiotemporal signatures within individual action classes remain identical. Secondly, action sequences are not complex, although this trend is starting to increase with the introduction of datasets like Something-Something. However, overall, and as a consequence of the first shortcoming, the datasets do not require video understanding models to reason at a deeper level: \eg, trying to draw logical inferences about the actors’ goals (what they are trying to do, which is beyond current fixed action set classification) by piecing together contextual (including human-object interactions) and human-movement cues as the video progresses, to determine whether the actor/s were able to accomplish what they set out to do. 

This raises a question as to, if the current video/action understanding systems make an attempt to really understand the action, or if they limit themselves to identifying spatiotemporal patterns. We believe that video understanding comes down to a pattern recognition problem. We are not conveying that action recognition is not needed or is unimportant, rather we view action recognition as a very important initial task. However, solely focusing on developing in the direction of action recognition might be limiting in nature. Therefore, in order to encourage action understanding systems to gain a deeper level understanding of human actions, we slightly redefine the task of action recognition as it currently stands. Instead of differentiating among action classes, we propose to repurpose the task of (action) recognition to differentiating between the concepts of winning and failing. Winning can be defined as completing a task that the human set out to do, while failing can be defined when human is not able to complete the task. For example, successfully flipping a cup, putting a basketball in the basket, or being able to walk on one's hands, \etc are considered as wins, while not successfully flipping a cup, throwing a basketball that does not go into the basket, or trying to walk on one's hands but instead falling over, are considered as fails. As one can imagine, \& see in Fig. \ref{fig:ar_vs_ar}, that the spatiotemporal signatures within the samples are very different, yet represent the same concepts. 

Humans, even as young as a year old, are able to infer and/or reason about the goals of others' actions from experience, contextual cues, kinematic cues, \etc \cite{woodward2014mirroring, biro2007infants, sommerville2005pulling, woodward2000twelve, falck2006infants}. They are able to perceive the difficulty of a task/action \cite{fitts1954information, grosjean2007fitts, eskenazi2009role}, and tend to put more value on actions that are more difficult or require perceivably more effort \cite{effortheuristic}. In fact, in Olympic events like diving and gymnastic vaulting, the scores are directly proportional to the degree of difficulty of the actions. It has been shown that the degree of difficulty can be measured from videos \cite{ltsoe, li2018scoringnet}. Therefore, when observing competitive scenarios (where participants are trying to gain the maximum score), even when the game rules, or a task descriptions are not explicitly intimated to humans, they may still be able to figure them out by reasoning about what action sequence might be more difficult to execute, and consequently, decide if an action instance was a win or fail. We aim to give our models this kind of deeper understanding through learning general notions of win and fail actions in videos in simplified setting.

To facilitate our novel task, we introduce a novel win-fail action understanding dataset. Our newly introduced dataset includes samples from the following categories: general stunts, win/fail internet videos, trick-shots, \& party games. It could be argued that it might not be possible for our networks to identify standalone wins or standalone fails. Therefore, we choose a comparative approach and, instead of compiling individual, unpaired samples from win and fail classes, we compile paired win-fail samples for each action instance. Currently, our dataset consists of 817 pairs. We provide further details regarding the dataset in Sec. \ref{sec:dataset}. We analyze our dataset in detail and provide baselines for future works in Sec. \ref{sec:exp}. Additionally, a novel video retrieval task is explored to characterize the general action understanding capabilities learned through the win-fail task.
%
%
%
\section{Related Work}
\noindent\textbf{Action recognition datasets}: Datasets can be divided into the following categories: 1) short-term temporal dynamics (UCF101 \cite{ucf101}, HMDB51 \cite{hmdb51}, Kinetics \cite{kinetics}), where actions can be classified from a single or very few frames, or even the background; 2) long-term temporal dynamics (Something-Something \cite{sthsth}, Diving48 \cite{diving48}, MTLAQA \cite{mtlaqa}, Epic-Kitchens \cite{epickitchens}, Jester \cite{jester}, \etc); 3) coarse-grained (UCF101, Kinetics, \etc); and 4) fine-grained (Diving48, MTLAQA, \etc). Unlike in coarse-grained action classification, actions in fine-grained classification category have very subtle differences between signature action patterns. 

Regardless of how we group current action datasets, the task ultimately remains same -- to learn the spatiotemporal signatures pertaining to each action class. Note that, longer temporal dynamics (for example, counting somersaults in \cite{diving48, mtlaqa} or differentiating between pulling and pushing in \cite{sthsth}) do not necessarily require deeper understanding of actions. From earlier action recognition days to the present, the focus, thus far, has been to increase the dataset size and increase the number of action classes. We believe this limits the models from gaining deeper understanding of what is happening in front of a camera. Therefore, instead, we focus on increasing the intra-class variance.

Our work can also be considered closer in spirit to \cite{girdhar2019cater}, which builds a video dataset with fully observable and controllable object and scene bias, and which truly requires spatiotemporal understanding in order to be solved. Our task and dataset are different in several ways, such as: ours is real world dataset; have humans performing actions/avtivities; objects in our dataset have purposes/meaning, \etc.\\

\noindent\textbf{Action recognition models}: Unlike in image recognition, where a decision is made based on a single image, in a video understanding task, modeling temporal relationships is crucial. Some of the earlier deep learning based action recognition works, which considered multiple frames to do recognition include works like \cite{sports1m, beyondsnippets, lrcn}.

TSN \cite{tsn} proposes a very simple, yet effective approach of sampling a few frames from the entire video and processing these frames individually to extract frame-wise features. Frame-level features are then combined  using an aggregation scheme to get video-level representation. The authors found averaging to work the best. Averaging is actually temporal order agnostic, which indicates that action recognition tasks on datasets like UCF101, HMDB51, and Kinetics do not really demand temporal modeling.

The introduction of datasets like Something-Something, in which temporal order of frames matter (\eg recognizing pushing vs. pulling something), motivated works like TRN \cite{trn}, which proposed a temporal reasoning module. While TRN worked by modeling/discovering temporal relations from extracted features, TSM \cite{tsm} aimed extracting temporal relations in the CNN backbone at a lower computational cost. Some works like \cite{slowfast, ltfb} propose approaches to combine short-term and long-term features.

Other works propose approaches that improve focus on the human actor, either by jointly estimating the pose of the actor  \cite{actemes, luvizon2d3d}, or by modeling the relationship between the actor and objects \cite{acrn}.

We employ developments in designing our models, and then compare them to see what works and what does not work on our dataset.\\ 

\noindent\textbf{AQA/skills assessment}: AQA \cite{parmar2016measuring, ltsoe, s3d, li2018end, xu_fs, maqa, mtlaqa, pan_aqa, musdl, hand_aqa, sardari2019view, zeng2020hybrid, lei2020learning, wang2020assessing} is another action analysis task, which involves quantifying \textit{how well} an action was performed. Similar in nature is the task of skills assessment \cite{doughty2018s, manipulation, proscons, parmar2021piano}. Like action recognition, the task still comes to learning and/or recognizing spatiotemporal patterns, although it is more fine-grained than action recognition. In addition to recognizing patterns, AQA/SA also involves valuation of those patterns.

In AQA, examples of these patterns include keeping legs straight in pike position, stable landing, tight form in tuck position, \etc; in SA, examples of these patterns include, not stretching tissues too much, handling them carefully, \etc. We do note that there is an association between AQA/SA and win-fail recognition, in that higher skills-levels are generally associated with wins, and lower skills-levels are associated with fails. However, our work is different from these works, in that while these works propose action-specific approaches,  our core idea is to increase intraclass variance among samples -- our dataset contains four different domains -- in order to encourage models to understand actions at a deeper level beyond surface-level pattern recognition; action sequences in our dataset are much more complex. Parmar \etal \cite{maqa} have proposed learning a single AQA model across multiple actions, resulting in more intra-class variance, but the goal of their work was to learn shared action quality elements more efficiently, while our goal in considering multiple domains is to gain an actual understanding of the actions.\\

\noindent\textbf{Visual concept learning}: We find that works by Binder \etal \cite{binder2014machine}, Zhou \etal  \cite{zhou2014learning}, and Chesneau \etal \cite{chesneau2017learning} are closest to ours. While \cite{binder2014machine, zhou2014learning} focus on recognizing more complex visual concepts, beyond objects in image domain, we introduce win-fail recognition in the video domain for deeper human action understanding. Chesneau \etal \cite{chesneau2017learning} address recognizing concepts like \textit{`Birthday Party,'} \textit{`Grooming an Animal,'} and \textit{`Unstuck a Vehicle'} in web videos. However, these concepts do not have large intra-class variance like ours, and are less complex and challenging than ours.
\section{Win-Fail Action Recognition Dataset}
\label{sec:dataset}
\begin{table}[]
\small
\centering
\begin{tabular}{@{}lcr@{}}
\toprule
\textbf{Action domain} & \textbf{No. of pairs} & \textbf{Avg. len. (\# fr.; s)} \\ \midrule
General stunts         & 122                       & 96; 3.83                    \\
Internet wins-fails    & 258                       & 112; 4.46                    \\
Trickshots             & 135                       & 66; 2.62 \\
Party games            & 302                       & 62; 2.48                   \\ \midrule
Overall				   & 817					   & 84; 3.33\\
\bottomrule
\end{tabular}
\caption{\textbf{Dataset details.}}
\label{tab:dataset}
\end{table}
\begin{figure*}
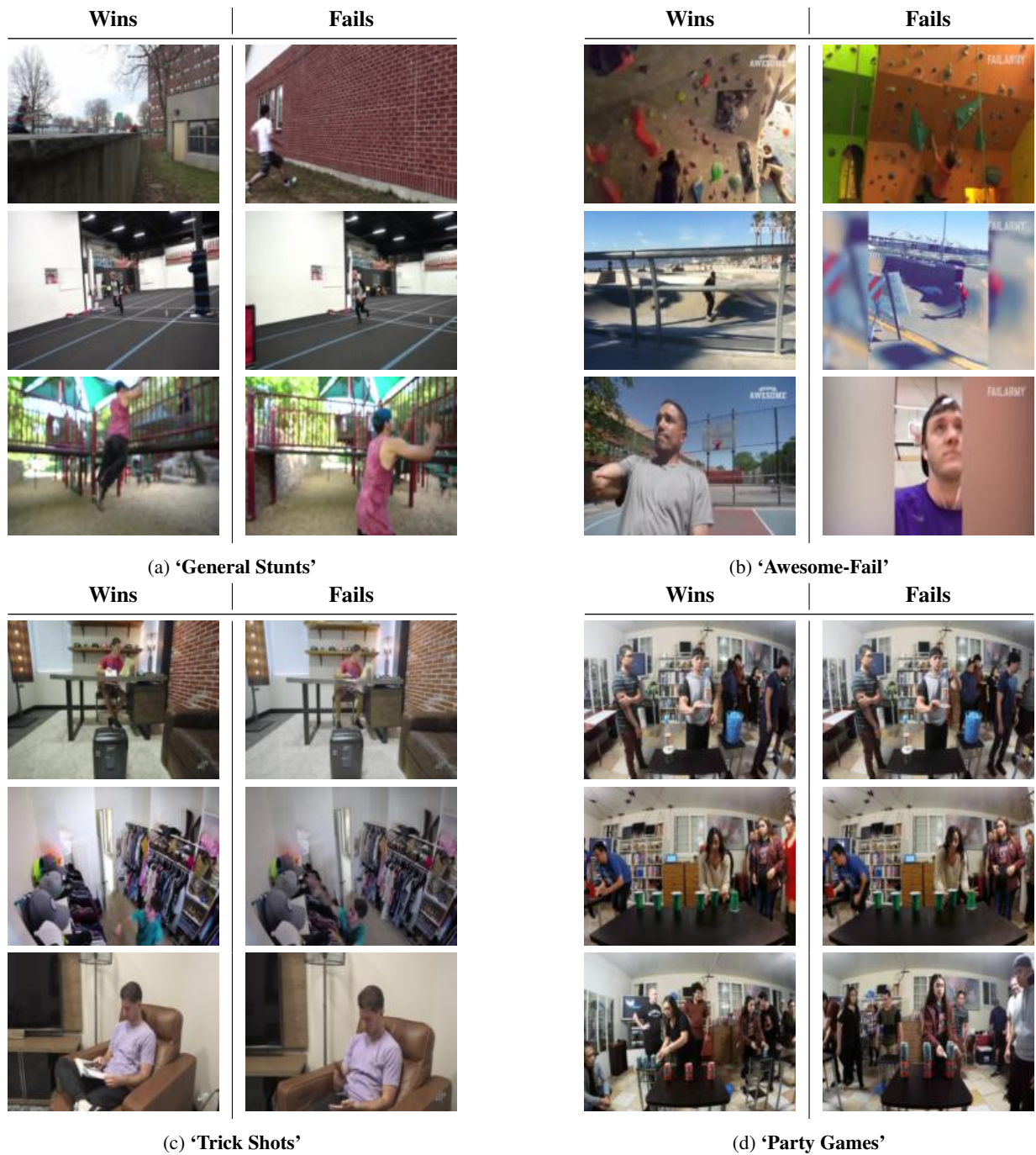

\begin{subfigure}[t]{\columnwidth} 
\centering
\begin{tabular}{@{}c|c@{}}
\textbf{Wins} & \textbf{Fails} \\ \midrule
\animategraphics[loop,autoplay,poster=1,width=0.4\columnwidth]{3}{Figs/Videos/Extracted_Samples/general/3/g/img_}{1}{16}    & \animategraphics[loop,autoplay,poster=1,width=0.4\columnwidth]{3}{Figs/Videos/Extracted_Samples/general/3/b/img_}{1}{16}     \\
\animategraphics[loop,autoplay,poster=1,width=0.4\columnwidth]{3}{Figs/Videos/Extracted_Samples/general/6/g/img_}{1}{16}    & \animategraphics[loop,autoplay,poster=1,width=0.4\columnwidth]{3}{Figs/Videos/Extracted_Samples/general/6/b/img_}{1}{16}     \\
\animategraphics[loop,autoplay,poster=1,width=0.4\columnwidth]{3}{Figs/Videos/Extracted_Samples/general/9/g/img_}{1}{16}    & \animategraphics[loop,autoplay,poster=1,width=0.4\columnwidth]{3}{Figs/Videos/Extracted_Samples/general/9/b/img_}{1}{16}     \\
\end{tabular}
\caption{\textbf{`General Stunts'}}
\label{fig:domain_general}
\end{subfigure}
%
%
\begin{subfigure}[t]{\columnwidth} 
\centering
\begin{tabular}{@{}c|c@{}}
\textbf{Wins} & \textbf{Fails} \\ \midrule
\animategraphics[loop,autoplay,poster=1,width=0.4\columnwidth]{3}{Figs/Videos/Extracted_Samples/awefail/3/g/img_}{1}{16}    & \animategraphics[loop,autoplay,poster=1,width=0.4\columnwidth]{3}{Figs/Videos/Extracted_Samples/awefail/3/b/img_}{1}{16}     \\
\animategraphics[loop,autoplay,poster=1,width=0.4\columnwidth]{3}{Figs/Videos/Extracted_Samples/awefail/6/g/img_}{1}{16}    & \animategraphics[loop,autoplay,poster=1,width=0.4\columnwidth]{3}{Figs/Videos/Extracted_Samples/awefail/6/b/img_}{1}{16}     \\
\animategraphics[loop,autoplay,poster=1,width=0.4\columnwidth]{3}{Figs/Videos/Extracted_Samples/awefail/9/g/img_}{1}{16}    & \animategraphics[loop,autoplay,poster=1,width=0.4\columnwidth]{3}{Figs/Videos/Extracted_Samples/awefail/9/b/img_}{1}{16}     \\
\end{tabular}
\caption{\textbf{`Awesome-Fail'}}
\label{fig:domain_awefail}
\end{subfigure}
%
%
\begin{subfigure}[t]{\columnwidth} 
\centering
\begin{tabular}{@{}c|c@{}}
\textbf{Wins} & \textbf{Fails} \\ \midrule
\animategraphics[loop,autoplay,poster=1,width=0.4\columnwidth]{3}{Figs/Videos/Extracted_Samples/trickshot/6/g/img_}{1}{16}    & \animategraphics[loop,autoplay,poster=1,width=0.4\columnwidth]{3}{Figs/Videos/Extracted_Samples/trickshot/6/b/img_}{1}{16}     \\
\animategraphics[loop,autoplay,poster=1,width=0.4\columnwidth]{3}{Figs/Videos/Extracted_Samples/trickshot/9/g/img_}{1}{16}    & \animategraphics[loop,autoplay,poster=1,width=0.4\columnwidth]{3}{Figs/Videos/Extracted_Samples/trickshot/9/b/img_}{1}{16}     \\
\animategraphics[loop,autoplay,poster=1,width=0.4\columnwidth]{3}{Figs/Videos/Extracted_Samples/trickshot/12/g/img_}{1}{16}    & \animategraphics[loop,autoplay,poster=1,width=0.4\columnwidth]{3}{Figs/Videos/Extracted_Samples/trickshot/12/b/img_}{1}{16}     \\
\end{tabular}
\caption{\textbf{`Trick Shots'}}
\label{fig:domain_trickshot}
\end{subfigure}
%
%
\hfill
\begin{subfigure}[t]{\columnwidth} 
\centering
\begin{tabular}{@{}c|c@{}}
\textbf{Wins} & \textbf{Fails} \\ \midrule
\animategraphics[loop,autoplay,poster=1,width=0.4\columnwidth]{3}{Figs/Videos/Extracted_Samples/partygames/3/g/img_}{1}{16}    & \animategraphics[loop,autoplay,poster=1,width=0.4\columnwidth]{3}{Figs/Videos/Extracted_Samples/partygames/3/b/img_}{1}{16}     \\
\animategraphics[loop,autoplay,poster=1,width=0.4\columnwidth]{3}{Figs/Videos/Extracted_Samples/partygames/9/g/img_}{1}{16}    & \animategraphics[loop,autoplay,poster=1,width=0.4\columnwidth]{3}{Figs/Videos/Extracted_Samples/partygames/9/b/img_}{1}{16}     \\
\animategraphics[loop,autoplay,poster=1,width=0.4\columnwidth]{3}{Figs/Videos/Extracted_Samples/partygames/5/g/img_}{1}{16}    & \animategraphics[loop,autoplay,poster=1,width=0.4\columnwidth]{3}{Figs/Videos/Extracted_Samples/partygames/5/b/img_}{1}{16}     \\
\end{tabular}
\caption{\textbf{`Party Games'}}
\label{fig:domain_partygames}
\end{subfigure}
\caption{\textbf{Examples of pairwise samples from various domains in our dataset}.}
\end{figure*}
To address the previously mentioned limitations of current action recognition datasets and to facilitate our new task of win-fail action recognition task, we introduce a novel Win-Fail action recognition dataset.  The Win-Fail dataset has the following characteristics: 1) a large variance in the structure of the task (both in action and context) and in the semantic definitions of wins and fails across samples; and 2) action sequences that are complex but at the same time win/fail recognition task is feasible.  It is possible to identify winning/failing through reasoning on human movements and context (including actor-object interactions, \etc), without requiring external knowledge. For example, we do not include games of chess in our dataset since it requires knowledge of game mechanics. Since identifying wins and fails in standalone fashion may be overly difficult, we collect paired win and fail samples: \ie for every win action instance, we have provided a fail version of that action instance. We collected data samples from following domains:

\begin{enumerate}
\item{\textbf{General Stunts (GS):}}
Actions from this domain resemble stunts similar to those seen in movies or arbitrarily choreographed stunts. To collect data samples from this domain, we made use of paired compilations released by the stunt artists themselves. In these paired compilations, they include and specifically indicate their failed and successful attempts at various stunt routines. Failures can be attributed to factors like: miscalculation in placement of limbs, imbalance, erroneous landing, not able to securely grip handles, etc. In the samples from this domain, people can be seen working/interacting with large objects like truck tires, foam plyo boxes, chutes, ladders, etc. Action sequences are mainly comprised of a single actor. Examples presented in Fig. \ref{fig:domain_general}.

\item{\textbf{Internet Wins-Fails (IWF):}}
This is a popular category of videos on YouTube, where people attempt to do all sorts of things like walking on their hands, pole dancing, cycling at high speed through forests, skateboarding, hulahooping, etc. We collected pairs of wins and fails of people trying to these things. Note that these types of compilations many times include cases where mishaps happen because of some other person's mistake or some objects' failure (breaking off, falling, etc.). We did not include those kinds of samples; we only include samples where the wins and fails are outcomes of the efforts of the person under consideration. 

We also did not included cases where the person was affected due to factors outside of their control. Examples of samples omitted are: a fan unexpectedly falls on a person working at their desk; or a pole becomes loose and comes off, while a pole dancer is using it. These kinds of videos may not require the actual understanding of actions, and may simply be classified by detecting the sudden increase in the video speed/motion magnitude.

Reasons for failure include errors in planning, aiming, perception/judgement, or execution; lack of skills/ability/strength (unlike in general stunts, actors are not always trained), etc. In this domain, actors can be seen interacting with large to medium sized objects such as skateboards, bicycles, hulahoops, skis, ropes, poles, exercise balls, etc. Action sequences mainly involves a single person. Examples presented in Fig. \ref{fig:domain_awefail}.

\item{\textbf{Trick-shots (TS):}}
This is another popular category of videos on YouTube, where people try to do things that are extremely difficult to perform. Examples include throwing compact disc into a very slim opening from a distance; generally, this requires many attempts before one succeeds. We compiled samples from failure footage and corresponding successful attempts. Objects used are medium to small sized such as, basketballs, spoons, bags, bottles, food items, cellphones, cups, etc. Unlike previous domains, the choreography of the action sequences in this domain is not limited to a single actor, and may involve the coordinated performance of two actors. Examples presented in Fig. \ref{fig:domain_trickshot}.

\item{\textbf{Party Games (PG):}}
Parties/social gatherings/get-togethers generally have a series of games. 
%
We collected pairs of failed and successful attempts at numerous indoor party games. We only selected games that are short ($\sim2.5$ secs), and where win/fail can be recognized. Actors can be seen interacting with small sized objects like cups, pencils, ping-pong balls, etc. Action sequences mainly involve a single actor, although unlike other domains, human spectators can be seen in the background standing steadily or moving. Examples shown in Fig. \ref{fig:domain_partygames}.
\end{enumerate}

\paragraph{Excluding telltale signs:} Sometimes, actors might be behaving joyously (after winning), or acting disappointed (after failing), which might give enough clue to the models to correctly predict win or fail without actually needing to understand the whole action sequence. There could be other signs as well. Therefore, during data collection, we made sure to not include any such signs in our action sequences.

All of the videos are of high resolution: $1280 \times 720$ pixels. Further specifications about our dataset are provided in Table \ref{tab:dataset}.
\section{Experiments}
\label{sec:exp}
In this section, we systematically determine the characteristics of our dataset, then provide baselines and suggestions for future efforts. 

\paragraph{Models}
\begin{figure}
    \centering
    \includegraphics[width=0.8\columnwidth]{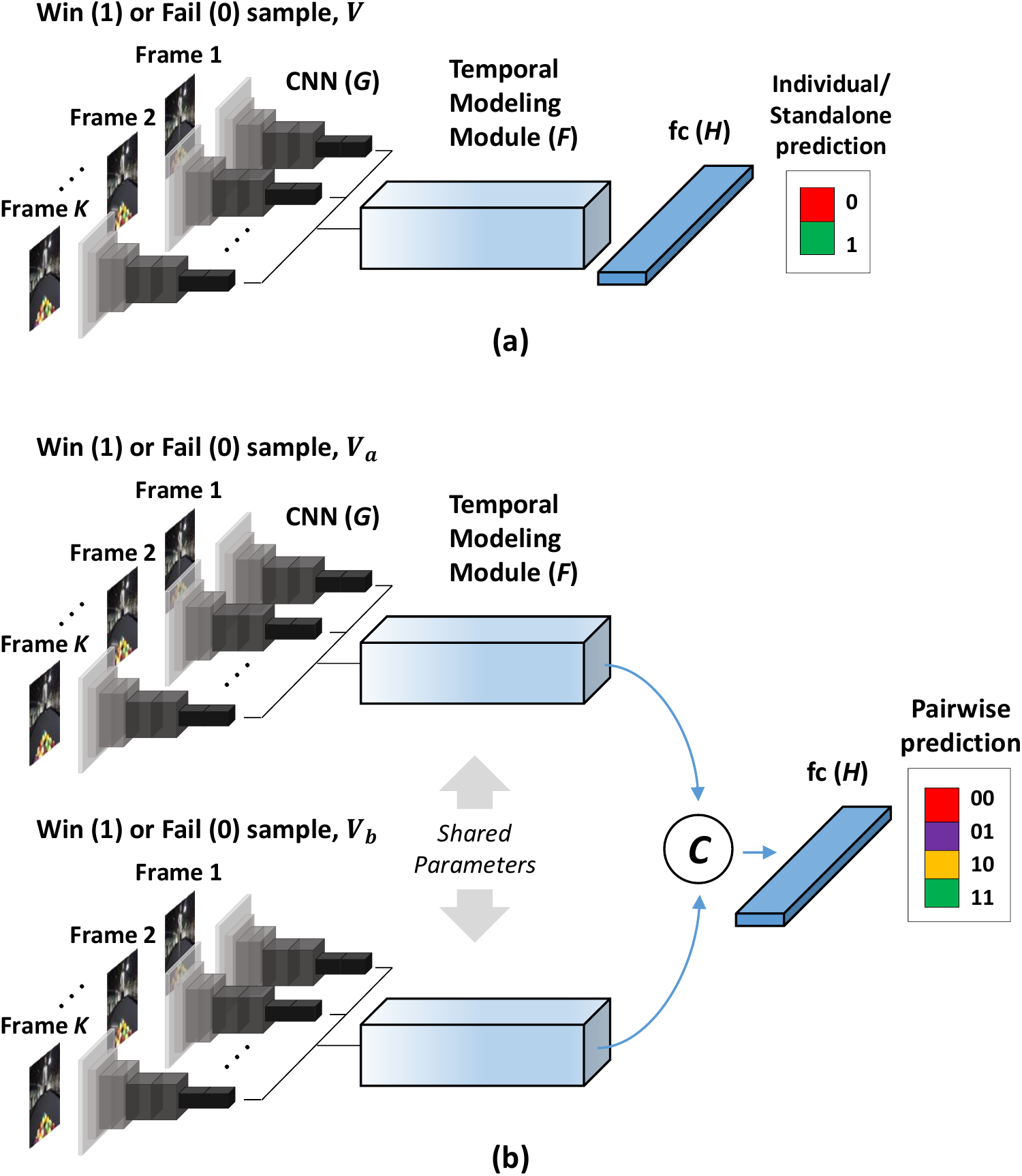}
    \caption{\textbf{Models}: Individual (a) and Pairwise (b). 00 and 11 in pairwise are equivalent to 0 and 1 in individual.}
    \label{fig:models_indi_pair}
\end{figure}
We used a CNN ($G$) to compute spatial features, followed by a temporal modeling module\footnote{In this paper, we alternately use terms: temporal modeling module, aggregation scheme, and consensus scheme.} (TMM) ($F$) to compute temporal relationships from spatial features from $G$. In particular, we considered temporal order agnostic (\textit{averaging} activations from $G$, and further processed through \texttt{fc} layers) and temporal order respecting (LSTM \cite{lstm} and TRN \cite{trn}) as our temporal models. 

We used cross-entropy loss, $\mathcal{L}$ as the objective function to train the networks. Let $x_{i}$ and $y_{i}$ be the predicted and ground-truth labels, then,
\begin{equation}
\label{eq:xentropy_loss}
 \mathcal{L} = -\frac{1}{N} \sum_{i=1}^{N} y_{i}logx_{i}
\end{equation}

We experimented with the following two approaches for the win-fail action recognition task:
\begin{enumerate}
\item{\textbf{Individual/standalone analysis:}} this is identical to any typical image/action classification, where we process the images/video-clip through a network, and it predicts the class (win or fail in our case). The model is illustrated in Fig. \ref{fig:models_indi_pair} (a). $x_{i} = H(F(G(V)))$, where, $V$ is input video frames in the case $G$ is a 2D-CNN, or it is video clips if $G$ is a 3D-CNN; and $H$ represents a linear layer. This is a binary classification problem: $x_{i}, y_{i} \in \{0, 1\}$.
\item{\textbf{Pairwise analysis:}} In this approach, we are able to leverage the pairwise nature of our dataset using a siamese setup, as shown in Fig. \ref{fig:models_indi_pair} (b). Let $V_{a}$ and $V_{b}$ represent two input videos, then, $x_{i} = H(C(F(G(V_{a})), F(G(V_{b}))))$, where, $C$ is the concatenation operation. We allow $V_{a} = V_{b}$ to incorporate the individual/standalone analysis of samples. We see pairwise loss as an aid to the learning process. In all, pairwise analysis is a four-way classification problem: $x_{i}, y_{i} \in \{00, 01, 10, 11\}$.
\end{enumerate}
%
%
\paragraph{Implementation details:} We used PyTorch \cite{pytorch} to implement all of our models. We used 2D ResNet-18 \cite{resnet} pretrained on ImageNet \cite{imagenet} as our CNN, unless mentioned otherwise. We trained all of our models for 100 epochs using ADAM \cite{adam} as our optimizer, with a learning rate of 1e-4, and a batchsize of 5. This also helped in keeping hyperparameter tuning to a minimum. We used a LSTM module with a hidden state of size 256. For a fair comparison with LSTM, for the \textit{averaging} case, we further add fully-connected layers after the averaging operation. Unless specified otherwise, we uniformly sampled 16 frames from entire video sample sequences and employed our pairwise approach. We resized all of the videos to a resolution of $320 \times 240$ pixels, and applied center cropping ($224 \times 224$ pixels); during the training phase, we also applied horizontal flipping. Center-cropping also removes branding/watermarking, which may give out the win/fail class, and may allow the network to take shortcuts. We will make our codebase publicly available.
\paragraph{Metric:} Unless specified otherwise, we report overall accuracy in percentage.
%
%
\subsection{Task feasibility, split ratios and aggregation schemes}
\label{sec:exp_splitrat_aggre}
\begin{table}[]
\small
\centering
\begin{subfigure}[t]{\columnwidth}
\small
\centering
\begin{tabular}{@{}lcccc@{}}
\toprule
\multirow{2}{*}{\textbf{Temporal model}} & \multicolumn{4}{c}{\textbf{Train:Test ratios}} \\ 
                                & \textbf{10:90}   & \textbf{30:70}   & \textbf{50:50}   & \textbf{70:30 } \\ \midrule
AVG.                            & 44.63   & 61.76   & 69.01   & 71.33  \\
LSTM                            & \textbf{48.10}   & \textbf{68.88}   & \textbf{72.56}   & \textbf{77.04 } \\ \bottomrule
\end{tabular}
\caption{}
\label{tab:splitrat_aggre_a}
\end{subfigure}
\begin{subfigure}[t]{0.4\linewidth}
\small
\centering
\setlength\tabcolsep{4pt}
\begin{tabular}{@{}cccc@{}}
\toprule
\textbf{GS} & \textbf{IWF} & \textbf{TS} & \textbf{PG} \\ \midrule
+13.00       & +9.75         & +0.25        & +5.50        \\ \bottomrule
\end{tabular}
\caption{}
\label{tab:splitrat_aggre_b}
\end{subfigure}
\hfill
\begin{subfigure}[t]{0.5\linewidth}
\small
\centering
\setlength\tabcolsep{4pt}
\begin{tabular}{@{}cccc@{}}
\toprule
\textbf{00} & \textbf{01} & \textbf{10} & \textbf{11} \\ \midrule
84.79       & 61.88       & 61.54       & 67.31       \\ \bottomrule
\end{tabular}
\caption{}
\label{tab:splitrat_aggre_c}
\end{subfigure}
\caption{(a) \textbf{Split ratios and aggregation methods}; (b) domain-wise gains of LSTM over AVG for 30:70 split; (c) class-wise accuracy of LSTM for 30:70 split.}
\label{tab:splitrat_aggre}
\end{table}
First of all, we wanted to determine if our task is feasible. Secondly, video action recognition by definition is a task of spatiotemporal nature, which implies that, ideally, temporal order of frames/clips, and hence temporal modeling, plays a very important part. On current action datasets, averaging (which ignores temporal order) as the consensus scheme yields the best results \cite{tsn, sports1m, mtlaqa}. Although some works incorporate local, short-term motion cues using 3D-CNN, optical flow, etc., the demand for actual long-term temporal modeling from current datasets is still limited. In this experiment, we wanted to determine which temporal modeling scheme is better suited for our dataset: temporal-order agnostic (averaging) or temporal-order sensitive (LSTM).

In order to focus only on temporal modeling, we pretrained our CNN backbone on ImageNet and then froze it, which acts as a general spatial feature extractor. Then, we learned only the parameters of the temporal model, which takes in features from the CNN backbone. We have decoupled the spatial learning aspect from temporal modeling -- both temporal models are fed the same spatial features. 

Thirdly, we wanted to determine a good train:test split ratio for our dataset. For that, we considered various train:test ratios. We compared Averaging (AVG) vs. LSTM for various train:test split ratios.

For this experiment, we employed pairwise comparative approach. Results are shown in Table. \ref{tab:splitrat_aggre_a}. Random guessing would have an accuracy of 25\%, since a pairwise comparative approach is a four-way classification problem. Both models performed significantly better than random chance across all split ratios, which suggests that our task is feasible. We observed that LSTM outperforms AVG for all split ratios. We also note that the LSTM's gain over AVG increases as the training pool increases. LSTM performing better than AVG clearly indicates that our dataset demands actual temporal modeling from models. This is because, even with a comparative approach, various contextual and human-movement cues from start to finish need to be strung together in a sequential manner to infer about the actor's goal and determine whether they achieved it.

Noting the trade-off between split ratios and performance, we chose 30:70 as our optimal split, which was used for the rest of the experiments.
%
%
\subsection{Individual vs. Pairwise}
\label{sec:exp_indi_pair}
\begin{table}[]
\centering
\begin{tabular}{@{}lc@{}}
\toprule
\textbf{Method} & \textbf{Accuracy} \\ \midrule
Chance          & 50.00             \\ 
Individual       & 58.65             \\
Pairwise        & 76.05             \\ \bottomrule
\end{tabular}
\vspace{0.2cm}
\caption{\textbf{Individual vs. Pairwise Assessment.}}
\label{tab:indi_pair}
\end{table}
In this experiment, we aimed to determine the correct approach to a win-fail action understanding problem: individual/standalone analysis or a pairwise comparative approach.

We used the LSTM aggregation and the same settings as Experiment \ref{sec:exp_splitrat_aggre}. We compare the performances of individual and pairwise approaches in Table \ref{tab:indi_pair}. Note that individual assessment is actually built into our pairwise approach as well. In Table \ref{tab:indi_pair}, for the pairwise approach, we show the average accuracy of 00 and 11 (individual assessment), which is directly comparable to that of actual individual assessment. Since individual assessment is a two-way (win or fail) classification problem, a random guess would have an accuracy of 50\%. We observe that the individual assessment model has quite a poor performance. On other hand, by learning through pairwise comparison, the model was able learn in a much better way, and was able to perform significantly better, with an accuracy of 76.05\%. These results indicate that a pairwise comparative approach is much more suitable, at least for the model that we used. We suggest a comparative approach, but we also want to encourage future works to develop better standalone approaches. 

We could have altered the order of Experiments \ref{sec:exp_splitrat_aggre} and \ref{sec:exp_indi_pair}, but that would have resulted in performing an unnecessarily larger number of experiments. For the rest of the experiments, we use the LSTM based pairwise approach.
%
%
\subsection{Rate of sampling input frames}
\label{sec:exp_sampling_density}
\begin{table}[]
\small
\centering
\begin{subfigure}[]{0.4\linewidth}
\small
\centering
\begin{tabular}{@{}lcc@{}}
\toprule
\multirow{2}{*}{\textbf{\begin{tabular}[c]{@{}l@{}}\# frs.\\ \end{tabular}}} & \multicolumn{2}{c}{\textbf{Accuracy}} \\ 
                                                                                   & \textbf{Ours}     & \textbf{UCF101}   \\ \midrule
4                                                                                  & 60.97             & \textbf{61.12}    \\
8                                                                                  & 64.20             & 60.06             \\
16                                                                                 & \textbf{68.88}    & 57.60             \\
32                                                                                 & 67.66             & 58.00             \\ \bottomrule
\end{tabular}
\caption{}
\label{tab:sampling_density_a}
\end{subfigure}
\hfill
\begin{subfigure}[]{0.5\linewidth}
\centering
\includegraphics[width=\linewidth]{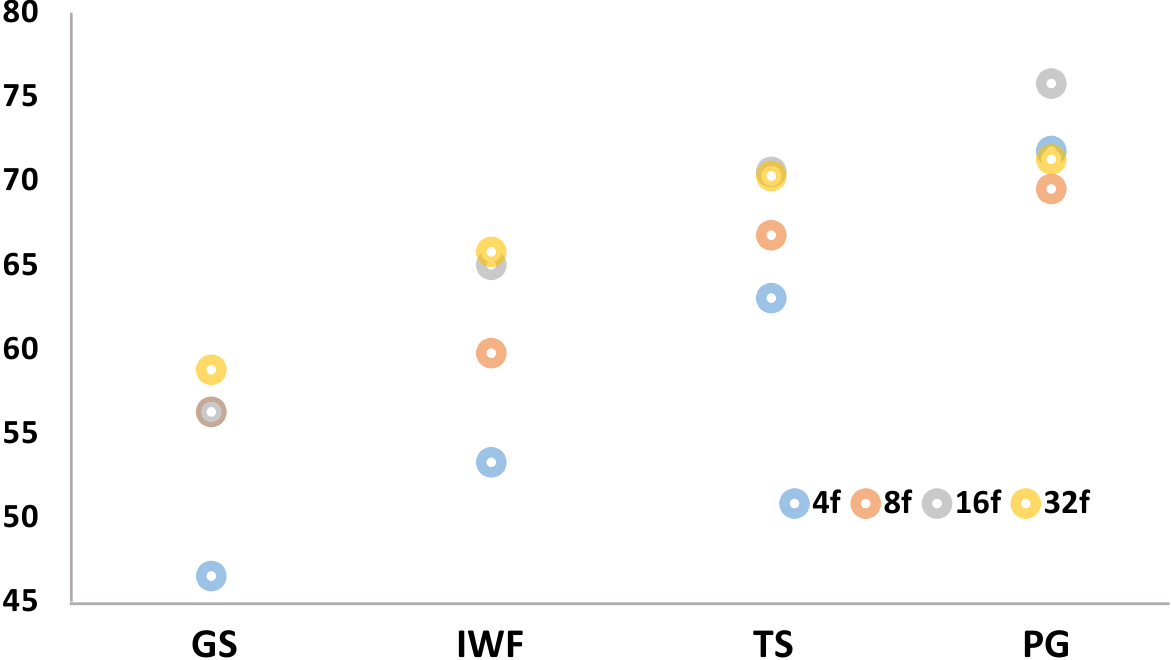} 
\caption{}
\label{tab:sampling_density_b}
\end{subfigure}
\caption{(a) \textbf{Effect of rate of sampling input frames}. (b) Effect on individual domains.}
\label{tab:sampling_density}
\end{table}
In this experiment, we studied the effect of the rate of sampling input frames. In particular, we considered sampling uniformly spaced 4, 8, 16, and 32 frames as input. We also conducted the same experiment on a typical action recognition dataset. The results are in Table \ref{tab:sampling_density_a}. We observed that on UCF101 dataset, the performance saturated with just 4 frames, while on ours action understanding dataset, it saturates at 16 frames. This indicates that intermediate frames and the cues/details in those are important. We also show the effect of varying the number of input frames across individual domains in Table \ref{tab:sampling_density_b}.
%
%
\subsection{Typical 3DCNN as feature extractor}
\label{sec:exp_2dvs3d}
\begin{table}[]
\small
\centering
\begin{subfigure}[t]{0.45\columnwidth}
\centering
\setlength\tabcolsep{4pt}
\begin{tabular}{lcc}
\toprule
\textbf{CNN} & \textbf{Trained on} & \textbf{Accu.} \\ \midrule
R18-2D         & ImageNet & \textbf{68.88}             \\
R18-3D         & Kinetics & 65.65             \\ \bottomrule
\end{tabular}
\caption{}
\label{tab:2dvs3d_a}
\end{subfigure}
\hfill
\begin{subfigure}[t]{0.45\columnwidth}
\centering
\setlength\tabcolsep{4pt}
\begin{tabular}{@{}cccc@{}}
\toprule
\textbf{GS} & \textbf{IWF} & \textbf{TS} & \textbf{PG} \\ \midrule
+5.75       & -1.25         & -2.00        & -8.75        \\ \bottomrule
\end{tabular}
\caption{}
\label{tab:2dvs3d_b}
\end{subfigure}
\caption{(a) \textbf{Backbone Choice: 2DCNN vs. 3DCNN}; (b) effect of using 3DCNN compared to 2DCNN across all domains.}
\label{tab:2dvs3d}
\end{table}

3DCNNs are known to extract much richer features than a 2DCNNs and as a result, obtain state-of-the results on action recognition tasks. In this experiment, we used a 3D counterpart (ResNet18-3D) of our 2DCNN. Both extract 512-dimensional features. ResNet18-3D extracts features from 16-frame clips. With 3DCNN as the backbone, we used 16 16-frame clips, in place of 16 frames, as input. Clips used with 3DCNN have a lower resolution ($112 \times 112$ pixels), as compared to that of frames used with 2DCNN ($224 \times 224$ pixels).

We compare the results in Table \ref{tab:2dvs3d}. Interestingly, we found that 3DCNN performed worse than 2DCNN. Only domain where 3D-CNN performed better is General Stunts, probably because Kinetics has similar action classes like Gymnastics. Potential reasons for poorer performance of 3D-CNN could be: 1) smaller resolution input might be hurting in our case because of the smaller sized objects and interactions involved with those; 2) actions patterns are different; 3) multiple humans present in the scene; or 4) ImageNet contains classes for many objects found in our dataset, while Kinetics does not.  We believe explicitly modeling/finetuning 3D-CNN for human-object interactions would be beneficial. 
%
%
\subsection{End-to-End learning}
\label{sec:exp_e2e_learn}
\begin{table}[]
\small
\centering
\begin{subfigure}[t]{0.45\linewidth}
\setlength\tabcolsep{4pt}
\begin{tabular}{lcc}
\toprule
\textbf{Training}    & \textbf{LSTM} & \textbf{TRN}   \\ \midrule
Only TMM & 68.88         & 71.24 \\
End-to-End           & \textbf{74.78}         & \textbf{75.74} \\ \bottomrule
\end{tabular}
\caption{}
\label{tab:e2elearn_a}
\end{subfigure}
\hfill
\begin{subfigure}[t]{0.45\linewidth}
\setlength\tabcolsep{4pt}
\begin{tabular}{@{}cccc@{}}
\toprule
\textbf{GS} & \textbf{IWF} & \textbf{TS} & \textbf{PG} \\ \midrule
+20.00       & +5.75         & +0.75        & +2.75        \\ \bottomrule
\end{tabular}
\caption{}
\label{tab:e2elearn_b}
\end{subfigure}
\begin{subfigure}[t]{0.45\linewidth}
\setlength\tabcolsep{4pt}
\begin{tabular}{@{}cccc@{}}
\toprule
\textbf{00} & \textbf{01} & \textbf{10} & \textbf{11} \\ \midrule
79.20       & 70.28       & 70.28       & 79.37       \\ \bottomrule
\end{tabular}
\caption{}
\label{tab:e2elearn_c}
\end{subfigure}
\hfill
\begin{subfigure}[t]{0.45\linewidth}
\setlength\tabcolsep{4pt}
\begin{tabular}{@{}cccc@{}}
\toprule
\textbf{00} & \textbf{01} & \textbf{10} & \textbf{11} \\ \midrule
79.90       & 73.60       & 72.40       & 77.10       \\ \bottomrule
\end{tabular}
\caption{}
\label{tab:e2elearn_d}
\end{subfigure}
\caption{(a) \textbf{End-to-End learning}. (b) Domain-wise improvements. (c, d) Class-wise accuracy of LSTM and TRN.}
\label{tab:e2elearn}
\end{table}
So far, we have used spatial features extracted using an off-the-shelf CNN. In this experiment, we sought to determine if there is a utility in jointly learning spatial and temporal representations. We unfroze the CNN backbone and optimized the network end-to-end. Then, we again finetuned only the temporal modeling module. The results are presented in Table \ref{tab:e2elearn_a}. We also evaluated a multiscale TRN (16f) baseline. We observed a significant boost in performance, indicating that the dataset requires spatial representation learning as well. We observe highest improvement in General Stunts, probably because ImageNet does not have people in unusual, convoluted poses, and hence, benefits a lot from finetuning. Furthermore,class-wise accuracies after end-to-end optimization are much more balanced w.r.t. 00 and 11 (compare Tables \ref{tab:e2elearn_c} and \ref{tab:splitrat_aggre_c}).

These performances also serve as baselines for future works. Since the end-to-end optimized model worked best, we use that in the rest of the experiments. For simplicity, we continue to use LSTM as our temporal modeling module.
%
%
\subsection{Importance of temporal order}
\label{sec:exp_scrambling}
\begin{table}[]
\small
    \begin{minipage}{.4\linewidth}
      \centering
\begin{tabular}{@{}lc@{}}
\toprule
\textbf{Shuffle type} & \textbf{Accu.} \\ \midrule
None                     & 74.78             \\
First $1/3$               & 74.16             \\
Middle $1/3$              & 74.26             \\
Last $1/3$                & 72.42             \\
Full                     & 61.19             \\ \bottomrule
\end{tabular}
\caption{\textbf{Effect of shuffling.}}
\label{tab:scrambling}
    \end{minipage}%
\hfill
    \begin{minipage}{.5\linewidth}
    \centering
\begin{tabular}{@{}lc@{}}
\toprule
\textbf{Observed seq.} & \textbf{Accu.} \\ \midrule
Full                         & 74.78             \\ \midrule
First $1/4$                      & 25.13             \\
First $1/2$                      & 35.62             \\
First $3/4$                      & 48.60             \\ \midrule
Last $1/4$                       & 23.47             \\
Last $1/2$                       & 28.93             \\
Last $3/4$                       & 52.58             \\ \bottomrule
\end{tabular}
\caption{\textbf{Partial observations.}}
\label{tab:incomplete_seq}
    \end{minipage} 
\end{table}
\begin{table*}[]
\small
\centering
\begin{tabular}{lllcllcllc}
\toprule
\multirow{2}{*}{\textbf{Model}} & \multicolumn{3}{c}{\textit{\textbf{Win-Fail $\rightarrow$ Fall-ADL}}}                 & \multicolumn{3}{c}{\textit{\textbf{Win-Fail $\rightarrow$ Baby Win-Fail}}}   & \multicolumn{3}{c}{\textit{\textbf{Win Fail $\rightarrow$ Animal Win-Fail}}}          \\ \cmidrule(l){2-4} \cmidrule(l){5-7} \cmidrule(l){8-10}
                                & \textbf{R@1}         & \textbf{R@5}          & \textbf{Sim $\Delta$} & \textbf{R@1}          & \textbf{R@5} & \textbf{Sim $\Delta$} & \textbf{R@1}          & \textbf{R@5}          & \textbf{Sim $\Delta$} \\ \cmidrule(r){1-1} \cmidrule(l){2-2} \cmidrule(l){3-3} \cmidrule(l){4-4} \cmidrule(l){5-5} \cmidrule(l){6-6} \cmidrule(l){7-7} \cmidrule(l){8-8} \cmidrule(l){9-9} \cmidrule(l){10-10}
AR              & 0.015                 & 0.073                 & 0.11           & 0.016                 & 0.094        & 0.06           & 0.042                 & 0.193                 & 0.05           \\
WFR            & \textbf{0.017 (\textcolor{ForestGreen}{$\uparrow$13\%})} & \textbf{0.088 (\textcolor{ForestGreen}{$\uparrow$20\%})} & \textbf{0.29}  & \textbf{0.018 (\textcolor{ForestGreen}{$\uparrow$13\%})} & 0.094  & \textbf{0.13}  & \textbf{0.057 (\textcolor{ForestGreen}{$\uparrow$36\%})} & \textbf{0.267 (\textcolor{ForestGreen}{$\uparrow$38\%})} & \textbf{0.44}  \\ \bottomrule
\end{tabular}
\caption{\textbf{Video retrieval results}. Higher is better. AR - action recognition model; WFR - win-fail recognition model.}
\label{tab:res_video_retrieval}
\end{table*}
In this experiment, we wanted to confirm if the temporal order from various parts of sequences matter. For that, in the testphase, we perturbed the temporal order of only a part of the sequence, while keeping the temporal order of the other parts of the sequence intact. In particular, we shuffled: 1) one-third of the sequence from the start (first 5 of the 16 sampled frames); 2) the middle one-third of the sequence (middle 5 frames); and 3) the last one-third of the sequence. Additionally, we considered shuffling the entire sequence.

The results are shown in Table \ref{tab:scrambling}. We observed that perturbing the temporal order in all the parts affected the performance negatively. Furthermore, we observed that impact of perturbation increased as we moved the focus of the shuffling towards the end of the sequence. Shuffling the entire sequence had the most negative impact. These observations support the hypothesis that our dataset demands/requires the algorithms/models to temporally model from the beginning to the end of the sequences.
%
%
\subsection{Where are the necessary cues?}
\label{sec:exp_incomplete_seq}
In this experiment, we wanted to probe if ``seeing" the entire sequence is needed, and/or if the model is able to make prediction just from a subsequence. During the testphase, we asked the model to classify based on partial sequences. Results are presented in Table \ref{tab:incomplete_seq}. We found that predictions based only on the first or last one-fourth sequence are equal to random guessing (25\% accuracy). We noticed that the accuracy of the model increased the further it observed the sequence, indicating that the necessary cues are present along the entire sequence.
%
%
\subsection{Video Retrieval}
\label{sec:exp_video_retrieval}
\begin{figure*}[]
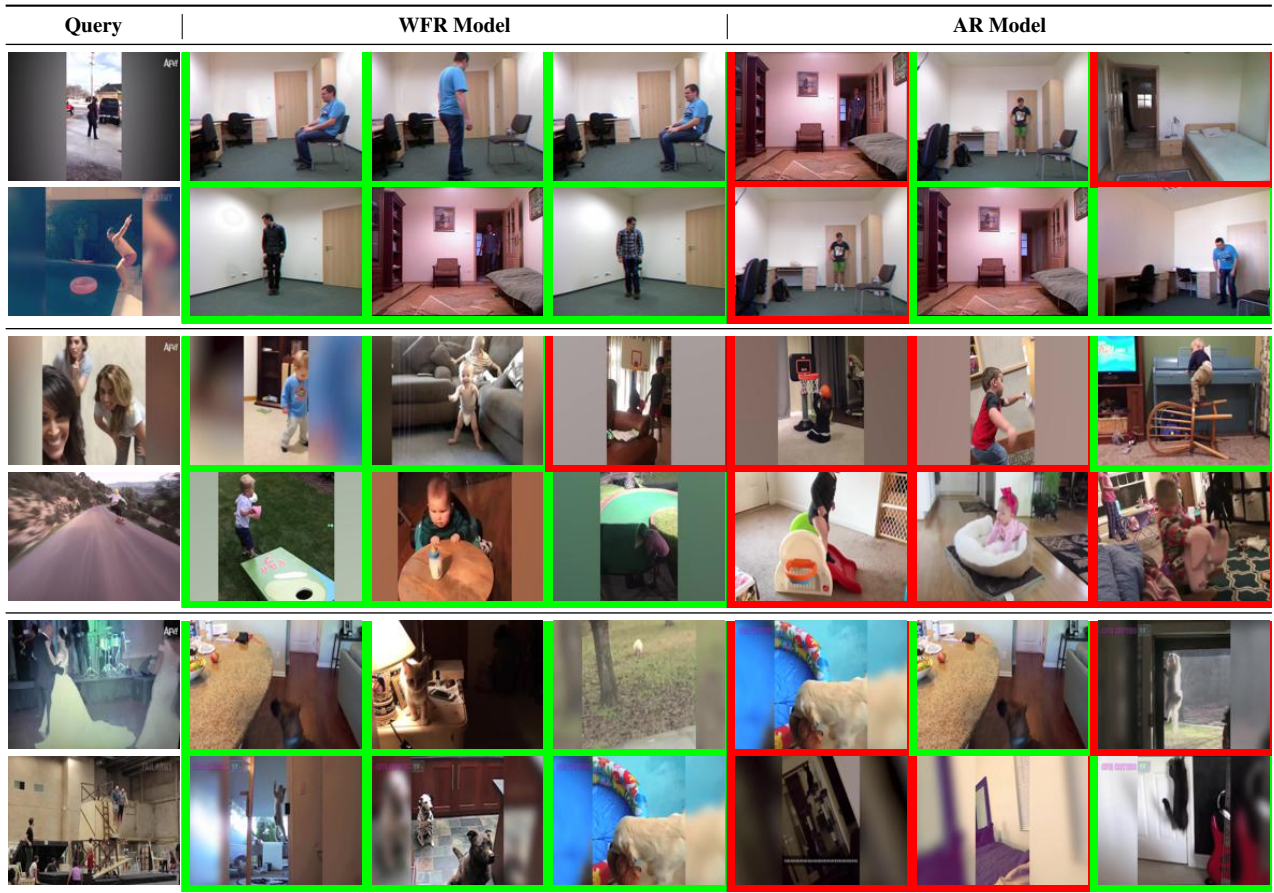

\footnotesize
\centering
\setlength\tabcolsep{1pt}
\begin{tabular}{c|ccc|ccc}
\toprule
\textbf{Query} & \multicolumn{3}{c|}{\textbf{WFR Model}} & \multicolumn{3}{c}{\textbf{AR Model}} \\ \midrule
%
%
\animategraphics[loop,autoplay,poster=1,width=0.13\textwidth]{3}{Figs/vr_res/new_fails/15/}{00}{15}
%
& \cellcolor{green} 
\animategraphics[loop,autoplay,poster=1,width=0.13\textwidth]{3}{Figs/vr_res/Fall/10/}{00}{15}  
& \cellcolor{green} \animategraphics[loop,autoplay,poster=1,width=0.13\textwidth]{3}{Figs/vr_res/Fall/7/}{00}{15}  
& \cellcolor{green}  \animategraphics[loop,autoplay,poster=1,width=0.13\textwidth]{3}{Figs/vr_res/Fall/8/}{00}{15}  
%
& \cellcolor{red}
\animategraphics[loop,autoplay,poster=1,width=0.13\textwidth]{3}{Figs/vr_res/ADL/3/}{00}{15}
& \cellcolor{green}  \animategraphics[loop,autoplay,poster=1,width=0.13\textwidth]{3}{Figs/vr_res/Fall/17/}{00}{15}
& \cellcolor{red} \animategraphics[loop,autoplay,poster=1,width=0.13\textwidth]{3}{Figs/vr_res/ADL/38/}{00}{15} \\ 
%
%
\animategraphics[loop,autoplay,poster=1,width=0.13\textwidth]{3}{Figs/vr_res/new_wins/4/}{00}{15}
%
& \cellcolor{green} 
\animategraphics[loop,autoplay,poster=1,width=0.13\textwidth]{3}{Figs/vr_res/ADL/27/}{00}{15}  
& \cellcolor{green} \animategraphics[loop,autoplay,poster=1,width=0.13\textwidth]{3}{Figs/vr_res/ADL/5/}{00}{15}  
& \cellcolor{green}  \animategraphics[loop,autoplay,poster=1,width=0.13\textwidth]{3}{Figs/vr_res/ADL/29/}{00}{15}  
%
& \cellcolor{red}
\animategraphics[loop,autoplay,poster=1,width=0.13\textwidth]{3}{Figs/vr_res/Fall/17/}{00}{15}
& \cellcolor{green}  \animategraphics[loop,autoplay,poster=1,width=0.13\textwidth]{3}{Figs/vr_res/ADL/2/}{00}{15}
& \cellcolor{green} \animategraphics[loop,autoplay,poster=1,width=0.13\textwidth]{3}{Figs/vr_res/ADL/34/}{00}{15} \\ \midrule
%
%
\animategraphics[loop,autoplay,poster=1,width=0.13\textwidth]{3}{Figs/vr_res/new_fails/1/}{00}{15}
%
& \cellcolor{green} 
\animategraphics[loop,autoplay,poster=1,width=0.13\textwidth]{3}{Figs/vr_res/baby_fails/10/}{00}{15}  
& \cellcolor{green} \animategraphics[loop,autoplay,poster=1,width=0.13\textwidth]{3}{Figs/vr_res/baby_fails/23/}{00}{15}  
& \cellcolor{red}  \animategraphics[loop,autoplay,poster=1,width=0.13\textwidth]{3}{Figs/vr_res/baby_wins/8/}{00}{15}  
%
& \cellcolor{red}
\animategraphics[loop,autoplay,poster=1,width=0.13\textwidth]{3}{Figs/vr_res/baby_wins/17/}{00}{15}
& \cellcolor{red}  \animategraphics[loop,autoplay,poster=1,width=0.13\textwidth]{3}{Figs/vr_res/baby_wins/6/}{00}{15}
& \cellcolor{green} \animategraphics[loop,autoplay,poster=1,width=0.13\textwidth]{3}{Figs/vr_res/baby_fails/8/}{00}{15} \\ 
%
%
\animategraphics[loop,autoplay,poster=1,width=0.13\textwidth]{3}{Figs/vr_res/new_wins/11/}{00}{15}
%
& \cellcolor{green} 
\animategraphics[loop,autoplay,poster=1,width=0.13\textwidth]{3}{Figs/vr_res/baby_wins/3/}{00}{15}  
& \cellcolor{green} \animategraphics[loop,autoplay,poster=1,width=0.13\textwidth]{3}{Figs/vr_res/baby_wins/11/}{00}{15}  
& \cellcolor{green}  \animategraphics[loop,autoplay,poster=1,width=0.13\textwidth]{3}{Figs/vr_res/baby_wins/9/}{00}{15}  
%
& \cellcolor{red}
\animategraphics[loop,autoplay,poster=1,width=0.13\textwidth]{3}{Figs/vr_res/baby_fails/9/}{00}{15}
& \cellcolor{red}  \animategraphics[loop,autoplay,poster=1,width=0.13\textwidth]{3}{Figs/vr_res/baby_fails/6/}{00}{15}
& \cellcolor{red} \animategraphics[loop,autoplay,poster=1,width=0.13\textwidth]{3}{Figs/vr_res/baby_fails/25/}{00}{15} \\ \midrule 
%
%
\animategraphics[loop,autoplay,poster=1,width=0.13\textwidth]{3}{Figs/vr_res/new_fails/16/}{00}{15}
%
& \cellcolor{green} 
\animategraphics[loop,autoplay,poster=1,width=0.13\textwidth]{3}{Figs/vr_res/animal_fails_1/2/}{00}{15}  
& \cellcolor{green} \animategraphics[loop,autoplay,poster=1,width=0.13\textwidth]{3}{Figs/vr_res/animal_fails_1/1/}{00}{15}  
& \cellcolor{green}  \animategraphics[loop,autoplay,poster=1,width=0.13\textwidth]{3}{Figs/vr_res/animal_fails_1/4/}{00}{15}  
%
& \cellcolor{red}
\animategraphics[loop,autoplay,poster=1,width=0.13\textwidth]{3}{Figs/vr_res/animal_wins_1/9/}{00}{15}
& \cellcolor{green}  \animategraphics[loop,autoplay,poster=1,width=0.13\textwidth]{3}{Figs/vr_res/animal_fails_1/2/}{00}{15}
& \cellcolor{red} \animategraphics[loop,autoplay,poster=1,width=0.13\textwidth]{3}{Figs/vr_res/animal_wins_1/12/}{00}{15} \\ 
%
%
\animategraphics[loop,autoplay,poster=1,width=0.13\textwidth]{3}{Figs/vr_res/new_wins/3/}{00}{15}
%
& \cellcolor{green} 
\animategraphics[loop,autoplay,poster=1,width=0.13\textwidth]{3}{Figs/vr_res/animal_wins_1/8/}{00}{15}  
& \cellcolor{green} \animategraphics[loop,autoplay,poster=1,width=0.13\textwidth]{3}{Figs/vr_res/animal_wins_1/10/}{00}{15}  
& \cellcolor{green}  \animategraphics[loop,autoplay,poster=1,width=0.13\textwidth]{3}{Figs/vr_res/animal_wins_1/9/}{00}{15}  
%
& \cellcolor{red}
\animategraphics[loop,autoplay,poster=1,width=0.13\textwidth]{3}{Figs/vr_res/animal_fails_1/13/}{00}{15}
& \cellcolor{red}  \animategraphics[loop,autoplay,poster=1,width=0.13\textwidth]{3}{Figs/vr_res/animal_fails_1/12/}{00}{15}
& \cellcolor{green} \animategraphics[loop,autoplay,poster=1,width=0.13\textwidth]{3}{Figs/vr_res/animal_wins_1/7/}{00}{15} \\ \bottomrule
\end{tabular}
\caption{\textbf{Qualitative results}. Odd and even rows show `Fails' and `Wins' as queries, respectively. First, second, and third two rows: ADL-Fall, Baby Win-Fail, Animal Win-Fail as databases. Red and green indicate relevant and irrelevant retrievals w.r.t. Win/Fail aspect.}
\label{fig:video_retrieval_qual_res}
\end{figure*}
To evaluate if our task yields deeper understanding, we devised a novel video retrieval experiment. We collected an additional, separate set of win and fail samples (from Internet Wins-Fails domain), which served as our queries. We also collect a set of baby and animal win-fails. Actors in our original win-fail dataset and the additional samples (queries) are adolescent and adult human beings. We use queries to retrieve videos from databases, where we changed the situations and actors. Particularly, we considered three different databases: 
\begin{enumerate}
\item Activities of Daily Living (ADL)-Fall: We used the dataset released by \cite{adlfall} as our database. ADL include activities such as sitting down, standing up, getting things from floor, \etc. Falls include person walking and falling down. This dataset was built to be used in monitoring elderly people. We consider ADL, and Fall as relevant to win and fail queries, respectively.
\item Win-Fail with Babies as actors: Fails in babies include babies trying to get up, crawl, walk but falling over since they have not yet acquired balance, falling while sitting due to lack of balance and control, \etc. Movements of babies are a lot more jittery compared to adults. Wins include climbing into their cradles successfully, throwing balls into baskets, passing through narrow spaces, \etc. We consider baby wins anf fails as relevant to adult win and fail queries, respectively.
\item Win-Fail with Animals as actors: Wins include animals being able to play ping-pong; shoot basket/pass a ball; open a door/window and get out/in; \etc. Fails include reaching out a take something, but falling; tumbling over while running; jumping but falling short; \etc. Note that animals have different structure and movements than humans. We consider animal win and fail as relevant to human win and fail queries, resp.
\end{enumerate}

Note that these queries and databases were not seen during training. We used cosine similarity as a similarity measure when retrieving, and recall at rank 1 and 5 (R@1, R@5) as metrics. We also noted average similarity difference from query to relevant and irrelevant samples in the databases (Sim $\Delta$), which would show sensitivity of features towards wins/fails in unseen domains. We considered model trained UCF101 for action recognition as our baseline. Results are summarized in Tab. \ref{tab:res_video_retrieval}. We found that win-fail recognition model outperformed action recognition model across all the databases. Moreover, the gap in performances increased when retrieving from animal database. We also observed that win-fail recognition model was more sensitive to win-fail aspect of the query, retrieved samples. These results also suggest application of WFR in areas like elderly and children safety monitoring. 

Qualitative results are presented in Fig. \ref{fig:video_retrieval_qual_res}. For brevity, in the following we refer to individual samples using respective row, and column numbers in Fig. \ref{fig:video_retrieval_qual_res}. We observe that AR model retrieves considerably on the basis of color (\eg: (R1,C6), (R1,C7), (R5,C5), (R5,C7)); low-level motion patterns (\eg. extended hand in (R3,C1)$\rightarrow$(R3,C6); sliding pattern (R4,C1)$\rightarrow$(R4,C5) -- skater smoothly gliding down the road is good, while the baby is smoothly falling down the slide is bad, jumping pattern (R6,C5),(R6,C6), (R6,C7)). Compared to that WFR model retrieves while maintaining meaning (\eg (R2,C2), (R2,C4) both exhibit compact body form of the diver in (R2,C1) necessary to sit on the chair and pass through swim rings; landing safely in (R6,C4) and (R6,C1); stunt involving multiple parties (R6,C1), (R6,C3); falling while reaching out (R5,C1),(R5,C2), (R5,C3)). Sometimes, like AR, WFR also puts more emphasis on motion patterns (\eg, (R6,C2)).
\section{Conclusion}
As a step towards true video/action understanding, we proposed the task of differentiating between the concepts of wins and fails. To facilitate our task, we also introduced a new dataset, which contains 817 pairs of successful and failed attempts at various activities from `General Stunts,' `Internet Wins-Fails,' `Trick Shots,' and `Party Games' domains. The action sequences in our dataset are not overly long, yet are complex. We systematically analyzed our dataset and found that: 1) our dataset requires true temporal modeling; 2) pairwise approach worked better than individual/standalone assessment; 3) details/cues important for understanding video/action are present in intermediate frames along the entire sequence; 4) better performance (as compared to a 2DCNN) of an off-the-shelf 3DCNN did not translate well to dataset/task; and 5) spatial modeling is equally important. While current action recognition methods worked well on our task/dataset, they still leaves a large gap to cover, indicating that there is significant opportunity to improve on this task.

{\small
\bibliographystyle{ieee_fullname}
\bibliography{egbib}

\begin{thebibliography}{10}\itemsep=-1pt

\bibitem{binder2014machine}
Alexander Binder, Wojciech Samek, Klaus-Robert M{\"u}ller, and Motoaki
  Kawanabe.
\newblock Machine learning for visual concept recognition and ranking for
  images.
\newblock In {\em Towards the Internet of Services: The THESEUS Research
  Program}, pages 211--223. Springer, 2014.

\bibitem{biro2007infants}
Szilvia Biro and Alan~M Leslie.
\newblock Infants’ perception of goal-directed actions: development through
  cue-based bootstrapping.
\newblock {\em Developmental science}, 10(3):379--398, 2007.

\bibitem{kinetics}
Joao Carreira and Andrew Zisserman.
\newblock Quo vadis, action recognition? a new model and the kinetics dataset.
\newblock In {\em proceedings of the IEEE Conference on Computer Vision and
  Pattern Recognition}, pages 6299--6308, 2017.

\bibitem{chesneau2017learning}
Nicolas Chesneau, Karteek Alahari, and Cordelia Schmid.
\newblock Learning from web videos for event classification.
\newblock {\em IEEE Transactions on Circuits and Systems for Video Technology},
  28(10):3019--3029, 2017.

\bibitem{epickitchens}
Dima Damen, Hazel Doughty, Giovanni~Maria Farinella, Sanja Fidler, Antonino
  Furnari, Evangelos Kazakos, Davide Moltisanti, Jonathan Munro, Toby Perrett,
  Will Price, and Michael Wray.
\newblock The epic-kitchens dataset: Collection, challenges and baselines.
\newblock {\em IEEE Transactions on Pattern Analysis and Machine Intelligence
  (TPAMI)}, 2020.

\bibitem{imagenet}
Jia Deng, Wei Dong, Richard Socher, Li-Jia Li, Kai Li, and Li Fei-Fei.
\newblock Imagenet: A large-scale hierarchical image database.
\newblock In {\em 2009 IEEE conference on computer vision and pattern
  recognition}, pages 248--255. Ieee, 2009.

\bibitem{lrcn}
Jeffrey Donahue, Lisa Anne~Hendricks, Sergio Guadarrama, Marcus Rohrbach,
  Subhashini Venugopalan, Kate Saenko, and Trevor Darrell.
\newblock Long-term recurrent convolutional networks for visual recognition and
  description.
\newblock In {\em Proceedings of the IEEE conference on computer vision and
  pattern recognition}, pages 2625--2634, 2015.

\bibitem{doughty2018s}
Hazel Doughty, Dima Damen, and Walterio Mayol-Cuevas.
\newblock Who's better? who's best? pairwise deep ranking for skill
  determination.
\newblock In {\em Proceedings of the IEEE Conference on Computer Vision and
  Pattern Recognition}, pages 6057--6066, 2018.

\bibitem{proscons}
Hazel Doughty, Walterio Mayol-Cuevas, and Dima Damen.
\newblock The pros and cons: Rank-aware temporal attention for skill
  determination in long videos.
\newblock In {\em proceedings of the IEEE Conference on Computer Vision and
  Pattern Recognition}, pages 7862--7871, 2019.

\bibitem{eskenazi2009role}
Terry Eskenazi, Marc Grosjean, Glyn~W Humphreys, and Guenther Knoblich.
\newblock The role of motor simulation in action perception: a
  neuropsychological case study.
\newblock {\em Psychological Research PRPF}, 73(4):477--485, 2009.

\bibitem{falck2006infants}
Terje Falck-Ytter, Gustaf Gredeb{\"a}ck, and Claes von Hofsten.
\newblock Infants predict other people's action goals.
\newblock {\em Nature neuroscience}, 9(7):878--879, 2006.

\bibitem{slowfast}
Christoph Feichtenhofer, Haoqi Fan, Jitendra Malik, and Kaiming He.
\newblock Slowfast networks for video recognition.
\newblock In {\em Proceedings of the IEEE international conference on computer
  vision}, pages 6202--6211, 2019.

\bibitem{fitts1954information}
Paul~M Fitts.
\newblock The information capacity of the human motor system in controlling the
  amplitude of movement.
\newblock {\em Journal of experimental psychology}, 47(6):381, 1954.

\bibitem{girdhar2019cater}
Rohit Girdhar and Deva Ramanan.
\newblock Cater: A diagnostic dataset for compositional actions and temporal
  reasoning.
\newblock {\em arXiv preprint arXiv:1910.04744}, 2019.

\bibitem{sthsth}
Raghav Goyal, Samira~Ebrahimi Kahou, Vincent Michalski, Joanna Materzynska,
  Susanne Westphal, Heuna Kim, Valentin Haenel, Ingo Fruend, Peter Yianilos,
  Moritz Mueller-Freitag, et~al.
\newblock The" something something" video database for learning and evaluating
  visual common sense.

\bibitem{grosjean2007fitts}
Marc Grosjean, Maggie Shiffrar, and G{\"u}nther Knoblich.
\newblock Fitts's law holds for action perception.
\newblock {\em Psychological Science}, 18(2):95--99, 2007.

\bibitem{resnet}
Kaiming He, Xiangyu Zhang, Shaoqing Ren, and Jian Sun.
\newblock Deep residual learning for image recognition.
\newblock In {\em Proceedings of the IEEE conference on computer vision and
  pattern recognition}, pages 770--778, 2016.

\bibitem{lstm}
Sepp Hochreiter and J{\"u}rgen Schmidhuber.
\newblock Long short-term memory.
\newblock {\em Neural computation}, 9(8):1735--1780, 1997.

\bibitem{sports1m}
Andrej Karpathy, George Toderici, Sanketh Shetty, Thomas Leung, Rahul
  Sukthankar, and Li Fei-Fei.
\newblock Large-scale video classification with convolutional neural networks.
\newblock In {\em CVPR}, 2014.

\bibitem{adam}
Diederik~P Kingma and Jimmy Ba.
\newblock Adam: A method for stochastic optimization.
\newblock {\em arXiv preprint arXiv:1412.6980}, 2014.

\bibitem{effortheuristic}
Justin Kruger, Derrick Wirtz, Leaf Van~Boven, and T~William Altermatt.
\newblock The effort heuristic.
\newblock {\em Journal of Experimental Social Psychology}, 40(1):91--98, 2004.

\bibitem{hmdb51}
Hildegard Kuehne, Hueihan Jhuang, Est{\'\i}baliz Garrote, Tomaso Poggio, and
  Thomas Serre.
\newblock Hmdb: a large video database for human motion recognition.
\newblock In {\em 2011 International Conference on Computer Vision}, pages
  2556--2563. IEEE, 2011.

\bibitem{adlfall}
Bogdan Kwolek and Michal Kepski.
\newblock Human fall detection on embedded platform using depth maps and
  wireless accelerometer.
\newblock {\em Computer methods and programs in biomedicine}, 117(3):489--501,
  2014.

\bibitem{lei2020learning}
Qing Lei, Hong-Bo Zhang, Ji-Xiang Du, Tsung-Chih Hsiao, and Chih-Cheng Chen.
\newblock Learning effective skeletal representations on rgb video for
  fine-grained human action quality assessment.
\newblock {\em Electronics}, 9(4):568, 2020.

\bibitem{li2018end}
Yongjun Li, Xiujuan Chai, and Xilin Chen.
\newblock End-to-end learning for action quality assessment.
\newblock In {\em Pacific Rim Conference on Multimedia}, pages 125--134.
  Springer, 2018.

\bibitem{li2018scoringnet}
Yongjun Li, Xiujuan Chai, and Xilin Chen.
\newblock Scoringnet: Learning key fragment for action quality assessment with
  ranking loss in skilled sports.
\newblock In {\em Asian Conference on Computer Vision}, pages 149--164.
  Springer, 2018.

\bibitem{diving48}
Yingwei Li, Yi Li, and Nuno Vasconcelos.
\newblock Resound: Towards action recognition without representation bias.
\newblock In {\em Proceedings of the European Conference on Computer Vision
  (ECCV)}, pages 513--528, 2018.

\bibitem{manipulation}
Zhenqiang Li, Yifei Huang, Minjie Cai, and Yoichi Sato.
\newblock Manipulation-skill assessment from videos with spatial attention
  network.
\newblock In {\em Proceedings of the IEEE International Conference on Computer
  Vision Workshops}, pages 0--0, 2019.

\bibitem{tsm}
Ji Lin, Chuang Gan, and Song Han.
\newblock Tsm: Temporal shift module for efficient video understanding.
\newblock In {\em Proceedings of the IEEE International Conference on Computer
  Vision}, pages 7083--7093, 2019.

\bibitem{luvizon2d3d}
Diogo~C Luvizon, David Picard, and Hedi Tabia.
\newblock 2d/3d pose estimation and action recognition using multitask deep
  learning.
\newblock In {\em Proceedings of the IEEE Conference on Computer Vision and
  Pattern Recognition}, pages 5137--5146, 2018.

\bibitem{jester}
Joanna Materzynska, Guillaume Berger, Ingo Bax, and Roland Memisevic.
\newblock The jester dataset: A large-scale video dataset of human gestures.
\newblock In {\em Proceedings of the IEEE International Conference on Computer
  Vision Workshops}, pages 0--0, 2019.

\bibitem{pan_aqa}
Jia-Hui Pan, Jibin Gao, and Wei-Shi Zheng.
\newblock Action assessment by joint relation graphs.
\newblock In {\em Proceedings of the IEEE International Conference on Computer
  Vision}, pages 6331--6340, 2019.

\bibitem{maqa}
Paritosh Parmar and Brendan Morris.
\newblock Action quality assessment across multiple actions.
\newblock In {\em 2019 IEEE Winter Conference on Applications of Computer
  Vision (WACV)}, pages 1468--1476. IEEE, 2019.

\bibitem{parmar2016measuring}
Paritosh Parmar and Brendan~Tran Morris.
\newblock Measuring the quality of exercises.
\newblock In {\em 2016 38th Annual International Conference of the IEEE
  Engineering in Medicine and Biology Society (EMBC)}, pages 2241--2244. IEEE,
  2016.

\bibitem{mtlaqa}
Paritosh Parmar and Brendan~Tran Morris.
\newblock What and how well you performed? a multitask learning approach to
  action quality assessment.
\newblock In {\em Proceedings of the IEEE Conference on Computer Vision and
  Pattern Recognition}, pages 304--313, 2019.

\bibitem{parmar2021piano}
Paritosh Parmar, Jaiden Reddy, and Brendan Morris.
\newblock Piano skills assessment.
\newblock {\em arXiv preprint arXiv:2101.04884}, 2021.

\bibitem{ltsoe}
Paritosh Parmar and Brendan Tran~Morris.
\newblock Learning to score olympic events.
\newblock In {\em proceedings of the IEEE Conference on Computer Vision and
  Pattern Recognition Workshops}, pages 20--28, 2017.

\bibitem{pytorch}
Adam Paszke, Sam Gross, Francisco Massa, Adam Lerer, James Bradbury, Gregory
  Chanan, Trevor Killeen, Zeming Lin, Natalia Gimelshein, Luca Antiga, Alban
  Desmaison, Andreas Kopf, Edward Yang, Zachary DeVito, Martin Raison, Alykhan
  Tejani, Sasank Chilamkurthy, Benoit Steiner, Lu Fang, Junjie Bai, and Soumith
  Chintala.
\newblock Pytorch: An imperative style, high-performance deep learning library.
\newblock In {\em Advances in Neural Information Processing Systems 32}, pages
  8026--8037. 2019.

\bibitem{sardari2019view}
Faegheh Sardari, Adeline Paiement, and Majid Mirmehdi.
\newblock View-invariant pose analysis for human movement assessment from rgb
  data.
\newblock In {\em International Conference on Image Analysis and Processing},
  pages 237--248. Springer, 2019.

\bibitem{sommerville2005pulling}
Jessica~A Sommerville and Amanda~L Woodward.
\newblock Pulling out the intentional structure of action: the relation between
  action processing and action production in infancy.
\newblock {\em Cognition}, 95(1):1--30, 2005.

\bibitem{ucf101}
Khurram Soomro, Amir~Roshan Zamir, and Mubarak Shah.
\newblock Ucf101: A dataset of 101 human actions classes from videos in the
  wild.
\newblock {\em arXiv preprint arXiv:1212.0402}, 2012.

\bibitem{acrn}
Chen Sun, Abhinav Shrivastava, Carl Vondrick, Kevin Murphy, Rahul Sukthankar,
  and Cordelia Schmid.
\newblock Actor-centric relation network.
\newblock In {\em Proceedings of the European Conference on Computer Vision
  (ECCV)}, September 2018.

\bibitem{musdl}
Yansong Tang, Zanlin Ni, Jiahuan Zhou, Danyang Zhang, Jiwen Lu, Ying Wu, and
  Jie Zhou.
\newblock Uncertainty-aware score distribution learning for action quality
  assessment.
\newblock In {\em Proceedings of the IEEE/CVF Conference on Computer Vision and
  Pattern Recognition}, pages 9839--9848, 2020.

\bibitem{wang2020assessing}
Jiahao Wang, Zhengyin Du, Annan Li, and Yunhong Wang.
\newblock Assessing action quality via attentive spatio-temporal convolutional
  networks.
\newblock In {\em Chinese Conference on Pattern Recognition and Computer Vision
  (PRCV)}, pages 3--16. Springer, 2020.

\bibitem{tsn}
Limin Wang, Yuanjun Xiong, Zhe Wang, Yu Qiao, Dahua Lin, Xiaoou Tang, and Luc
  Van~Gool.
\newblock Temporal segment networks: Towards good practices for deep action
  recognition.
\newblock In {\em European conference on computer vision}, pages 20--36.
  Springer, 2016.

\bibitem{hand_aqa}
Tianyu Wang, Minhao Jin, Jingying Wang, Yijie Wang, and Mian Li.
\newblock Towards a data-driven method for rgb video-based hand action quality
  assessment in real time.
\newblock In {\em Proceedings of the 35th Annual ACM Symposium on Applied
  Computing}, pages 2117--2120, 2020.

\bibitem{woodward2014mirroring}
Amanda~L Woodward and Sarah~A Gerson.
\newblock Mirroring and the development of action understanding.
\newblock {\em Philosophical Transactions of the Royal Society B: Biological
  Sciences}, 369(1644):20130181, 2014.

\bibitem{woodward2000twelve}
Amanda~L Woodward and Jessica~A Sommerville.
\newblock Twelve-month-old infants interpret action in context.
\newblock {\em Psychological Science}, 11(1):73--77, 2000.

\bibitem{ltfb}
Chao-Yuan Wu, Christoph Feichtenhofer, Haoqi Fan, Kaiming He, Philipp
  Krahenbuhl, and Ross Girshick.
\newblock Long-term feature banks for detailed video understanding.
\newblock In {\em Proceedings of the IEEE Conference on Computer Vision and
  Pattern Recognition}, pages 284--293, 2019.

\bibitem{s3d}
Xiang Xiang, Ye Tian, Austin Reiter, Gregory~D Hager, and Trac~D Tran.
\newblock S3d: Stacking segmental p3d for action quality assessment.
\newblock In {\em 2018 25th IEEE International Conference on Image Processing
  (ICIP)}, pages 928--932. IEEE, 2018.

\bibitem{xu_fs}
Chengming Xu, Yanwei Fu, Bing Zhang, Zitian Chen, Yu-Gang Jiang, and Xiangyang
  Xue.
\newblock Learning to score figure skating sport videos.
\newblock {\em IEEE Transactions on Circuits and Systems for Video Technology},
  2019.

\bibitem{beyondsnippets}
Joe Yue-Hei~Ng, Matthew Hausknecht, Sudheendra Vijayanarasimhan, Oriol Vinyals,
  Rajat Monga, and George Toderici.
\newblock Beyond short snippets: Deep networks for video classification.
\newblock In {\em Proceedings of the IEEE conference on computer vision and
  pattern recognition}, pages 4694--4702, 2015.

\bibitem{zeng2020hybrid}
Ling-An Zeng, Fa-Ting Hong, Wei-Shi Zheng, Qi-Zhi Yu, Wei Zeng, Yao-Wei Wang,
  and Jian-Huang Lai.
\newblock Hybrid dynamic-static context-aware attention network for action
  assessment in long videos.
\newblock In {\em Proceedings of the 28th ACM International Conference on
  Multimedia}, pages 2526--2534, 2020.

\bibitem{actemes}
Weiyu Zhang, Menglong Zhu, and Konstantinos~G Derpanis.
\newblock From actemes to action: A strongly-supervised representation for
  detailed action understanding.
\newblock In {\em Proceedings of the IEEE International Conference on Computer
  Vision}, pages 2248--2255, 2013.

\bibitem{trn}
Bolei Zhou, Alex Andonian, Aude Oliva, and Antonio Torralba.
\newblock Temporal relational reasoning in videos.
\newblock In {\em Proceedings of the European Conference on Computer Vision
  (ECCV)}, pages 803--818, 2018.

\bibitem{zhou2014learning}
Bolei Zhou, Agata Lapedriza, Jianxiong Xiao, Antonio Torralba, and Aude Oliva.
\newblock Learning deep features for scene recognition using places database.
\newblock In {\em Advances in neural information processing systems}, pages
  487--495, 2014.

\end{thebibliography}
}

\end{document}